\documentclass[letterpaper, 10 pt, conference]{ieeeconf}  

\IEEEoverridecommandlockouts                              

\usepackage{graphicx} 
\usepackage{epsfig} 
\usepackage{mathptmx} 
\usepackage{times} 
\usepackage{amsmath} 
\usepackage{amssymb}  
\usepackage{subfig}
\usepackage{fancyhdr}
\fancypagestyle{withfooter}{
  
  \fancyfoot[C]{\footnotesize Accepted to the IEEE ICRA Workshop on Field Robotics 2024}
}

\newcommand{\ccr}{CCR }

\newcommand{\xxx}{CropFollow++~}
\title{\LARGE \bf Lessons from Deploying CropFollow++: Under-Canopy Agricultural Navigation with Keypoints}

\author{Arun N. Sivakumar$^{1}$, Mateus V. Gasparino$^{1}$, Michael McGuire$^{2}$, Vitor A. H. Higuti$^{2}$, \\
M. Ugur Akcal$^{1}$ and Girish Chowdhary$^{1}$
\thanks{$^{1}$Field Robotics Engineering and Science Hub (FRESH), Illinois Autonomous Farm, University of Illinois at Urbana-Champaign (UIUC), IL, USA}%
\thanks{$^{2}$Earthsense Inc.}%
\thanks{This work was
supported in part by NSF STTR \#1951250, NSF NRI 2.0 NIFA \#2021-67021-33449, AIFARMS \#1024178, NSF-USDA COALESCE \#2021-67021-34418, USDA grants iCOVER(\#NR233A750004G066) and iFARM(\#2022-77038-37306).
}}

\begin{document}

\maketitle
\thispagestyle{withfooter}
\pagestyle{withfooter}

\begin{abstract}
We present a vision-based navigation system for under-canopy agricultural robots using semantic keypoints. Autonomous under-canopy navigation is challenging due to the tight spacing between the crop rows ($\sim 0.75$ m), degradation in RTK-GPS accuracy due to multipath error, and noise in LiDAR measurements from the excessive clutter. Our system, CropFollow++, introduces modular and interpretable perception architecture with a learned semantic keypoint representation. We deployed \xxx in multiple under-canopy cover crop planting robots on a large scale (25 km in total) in various field conditions and we discuss the key lessons learned from this. 

\end{abstract}

\section{Introduction}
\label{sec:intro}

Agricultural production faces major challenges due to rising costs and reduced supply of farm labor coupled with strong environmental concerns due to overuse of farm chemicals. Autonomous under-canopy robots have great potential to address these challenges by enabling plant-level monitoring and care. In particular, under-canopy robots that traverse the tight space between rows of crops ($\sim 0.75$ m) can enable various applications like high throughput phenotyping and crop monitoring, cover crop planting, precise weeding and spraying throughout the growing season which is not possible with over the canopy systems like tractors and drones \cite{mcallister2019agbots,uppalapati2020berry,r2018research}. A major bottleneck in deploying under-canopy robots in farms is the lack of robust, low-cost autonomous navigation solutions for these challenging environments.

Tractors and drones primarily depend on precise positioning from RTK-GPS for navigation in farms. RTK-GPS accuracy degrades due to multi-path errors under the plant canopy, even with an expensive RTK correction service subscription. LiDAR sensors are expensive and suffer from noise in measurements due to excessive clutter in these environments \cite{higuti2019under,sivakumar2021learned}. The lack of semantic information to distinguish the crop of interest from weeds or other objects limits LiDAR capabilities for this problem. Cameras offer rich information about the scene and are far lower in cost, and hence are a better alternative \cite{sivakumar2021learned,zhao2023approxcaliper}. Prior works in vision-based agricultural row following were primarily focused on using classical computer vision methods from the over-the-canopy views of tractors and large machinery \cite{wang2022applications,zhang1999agricultural,aastrand2005vision,ball2016vision,winterhalter2021localization}. Due to the significant clutter and occlusion commonly encountered and large variations in the appearance of the scene, vision-based navigation under the canopy is more challenging. 

We developed a vision-based navigation system called CropFollow++ that uses learned semantic key points as a representation for this under-canopy navigation problem. We deployed this semantic keypoint perception system on under-canopy robots developed for a novel application: Cover crop planting. The tests were conducted in various field conditions on multiple cover crop planting robots for 25~km. In this paper, we discuss the various failure modes observed and lessons learned from this large-scale deployment.

\begin{figure}%
    \centering
    {{\includegraphics[width=8.5cm]{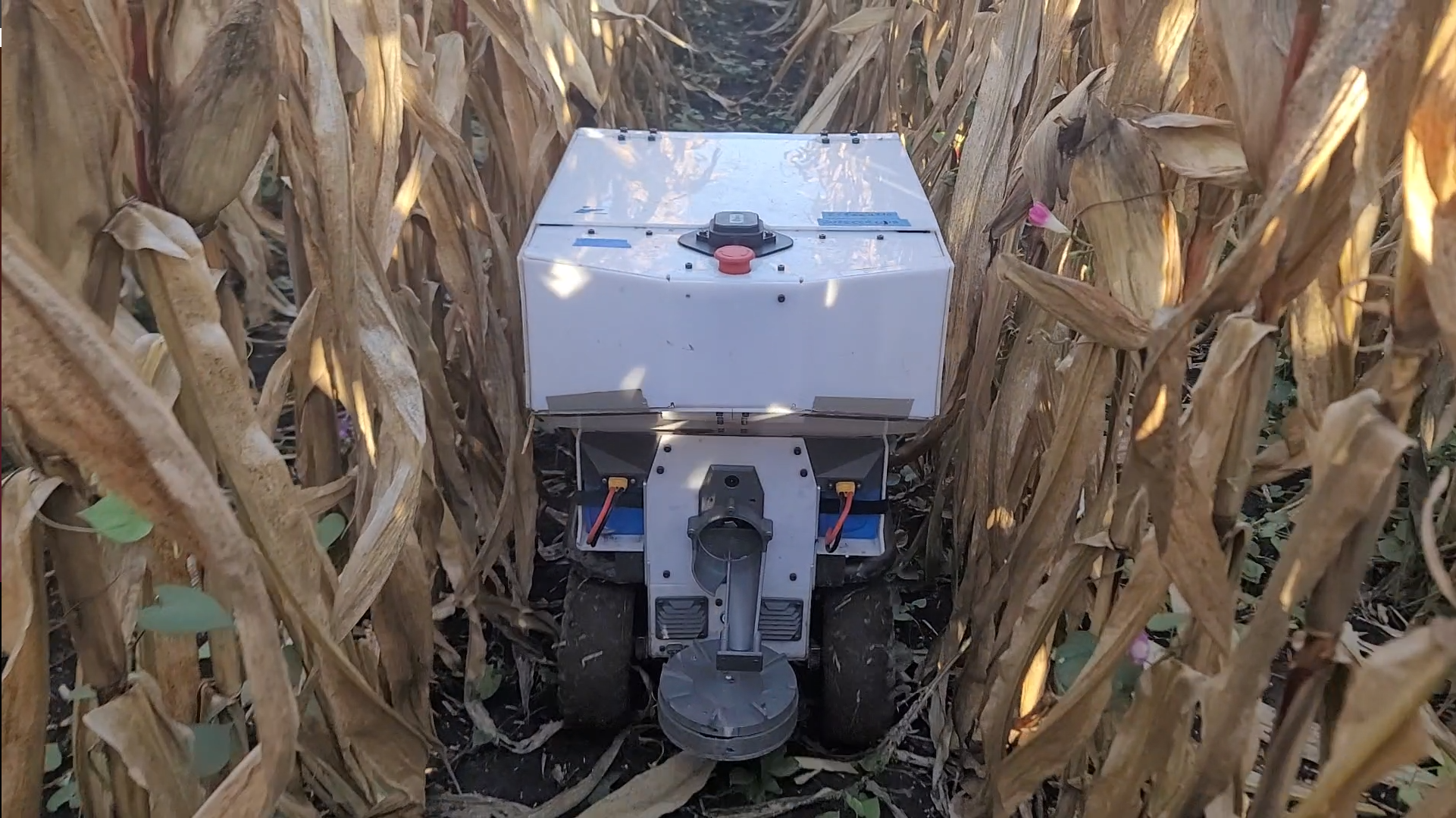} }}%
    \caption{Cover Crop Robot navigating through the minimal space available between the crop rows.}%
    \label{fig:example}
\end{figure}

\section{CropFollow++ Overview}
\label{sec:method}
\begin{figure*}
    \centering
    \includegraphics[width=0.95\linewidth]{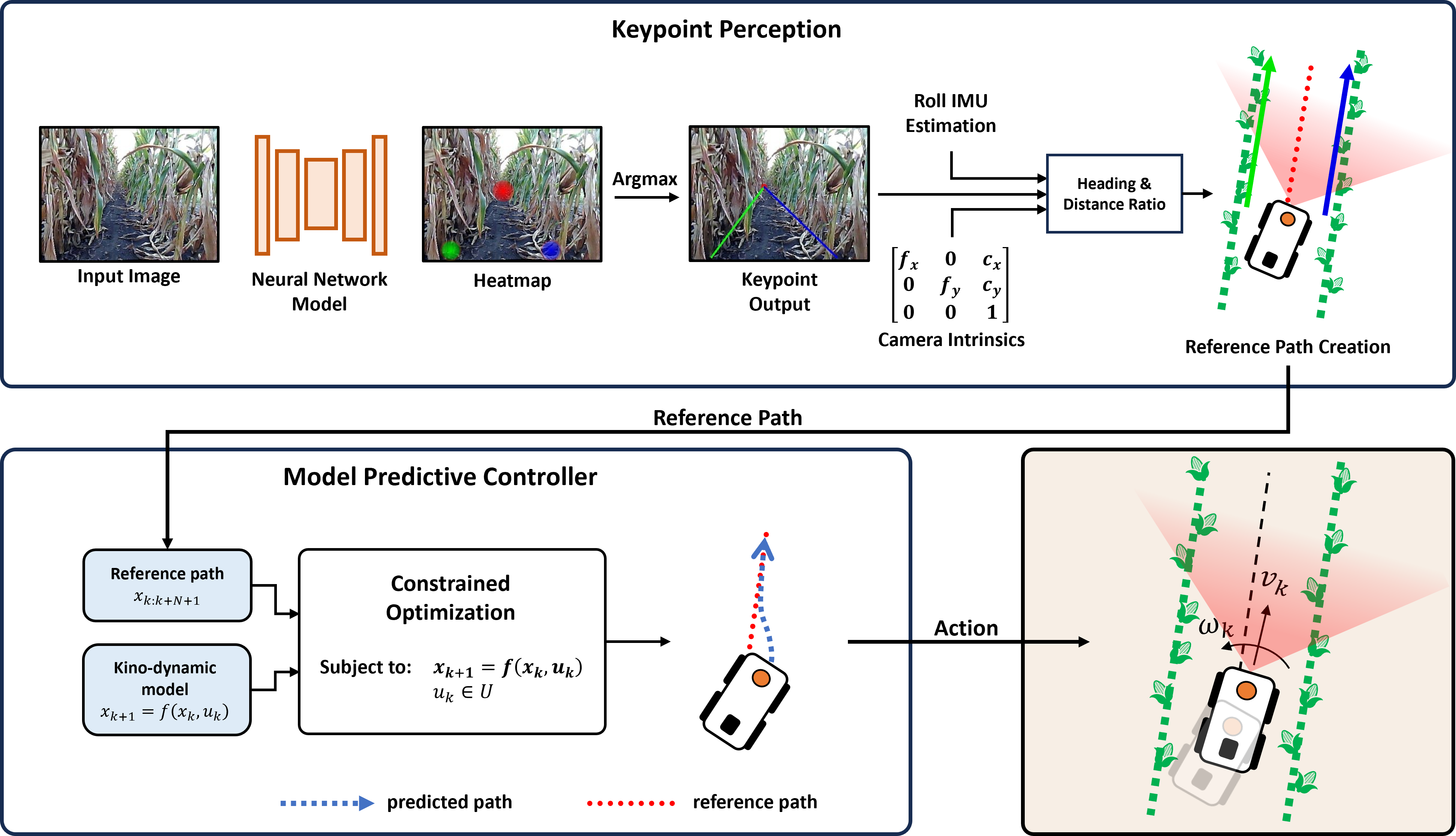}
    \caption{CropFollow++ overview. The camera RGB image is used as input to our neural network model that predicts keypoints to locate the crop rows. The keypoints are used to create a trajectory that is used as the reference for an MPC controller.}
\label{fig:overview}
\end{figure*}

Fig \ref{fig:overview} provides an overview of our proposed keypoint-based under-canopy navigation system. The perception system takes the RGB images from the low-cost camera on the robot as input and outputs three heatmaps. These heatmaps correspond to the three semantic keypoints of interest - vanishing point, left intercept point and right intercept point. Heading and lateral distance of the robot relative to the crop rows are calculated from the heatmaps using the camera intrinsics and roll estimate from the inertial measurement unit (IMU) if the heuristic-based confidence check is passed. A Model Predictive Controller (MPC) uses the calculated heading and lateral distance and solves a constrained cost optimization to find the optimal linear and angular velocity to track the reference. 

\section{Field Experiments}
\label{sec:results}
We deployed our proposed \xxx on multiple under-canopy cover crop planting robots over large distances and discuss the observed performance, common failure modes, and lessons learned.

\begin{figure}[ht!]
\centering
\includegraphics[width=8.5cm]{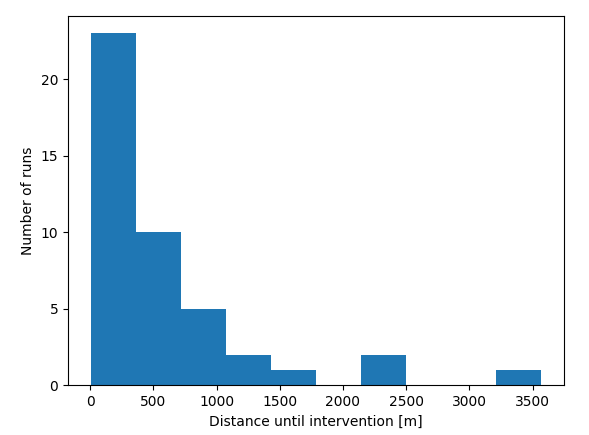}
\centering
\caption{We report the histogram of the distance traveled before intervention across all the autonomous runs from three CCRs. Though the majority of the runs are less than 250 m, we show three instances of autonomous runs with more than 2000 m.}
\label{fig:ccr_histogram}
\end{figure}

\begin{figure*}[ht!]
    \centering
    \includegraphics[trim=10 0 40 0,clip,width=1\textwidth]{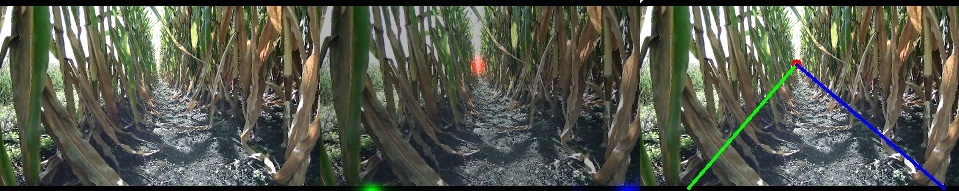}
    \includegraphics[trim=10 0 40 0,clip,width=1\textwidth]{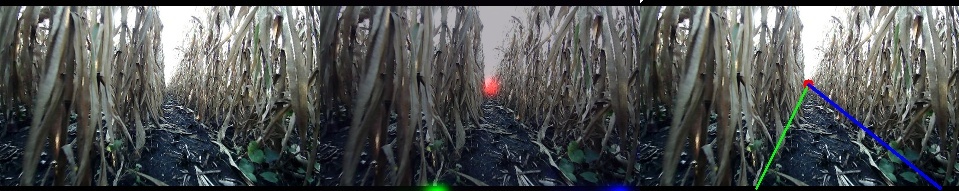}
    \includegraphics[trim=10 0 40 0,clip,width=1\textwidth]{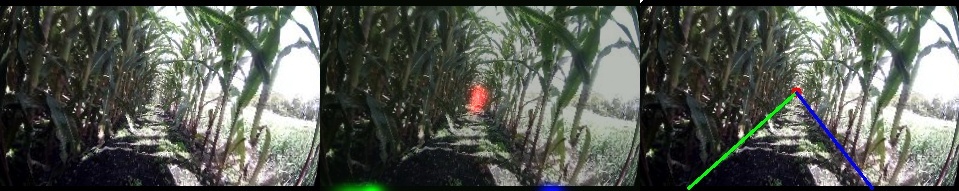}
    \caption{\textbf{Keypoint Detection in Action:} Here we show three randomly sampled images from \ccr deployment with keypoint detections. \textbf{Left:} Raw images from the CCR front camera. \textbf{Center:} Visualizations of the heatmaps for the vanishing point (red), the left point (green), and the right point (blue). \textbf{Right:} Visualizations of the keypoint outputs and the extracted vanishing lines, computed as the argmax of each heatmap.}
\label{fig:normal_keypoint_detection}
\end{figure*}

The EarthSense CCR Platform is designed for autonomous planting of cover crops under the corn canopy. It is equipped with four cameras. Three of the cameras are in the front, to provide redundancy against occlusion. One camera is in the rear to provide capacity for crash recovery. The CCR is 0.45~m wide, leaving only 0.15~m of space on each side of the robot in a standard 0.75~m corn row. To achieve long-range autonomy with minimal human interventions, we configured the CCRs to dynamically switch between a forward row follow mode and a reverse crash recovery mode \cite{gasparino2023cropnav}. We used a proprietary method based on odometry data developed by Earthsense Inc. to detect a crash. When a crash was detected, the robot would stop, and then execute the crash recovery procedure until row follow was ready to execute again. During row-follow, the CCR ran keypoint detection, heading estimation, and side distance estimation independently on the three front cameras. These outputs were fused and then fed into the waypoint generation and MPC controller. During crash recovery, the only difference was that the CCR ran keypoint detection, heading estimation, and side distance estimation only on the rear camera, and the waypoint generator provided a backward target path. Fusing the predictions from multiple cameras improved the robustness of the system.

We did extensive autonomy runs on three CCRs, many of which were done while actively carrying a heavy cover crop payload and planting cover crops.  We ran our experiments at 0.9~m/s for row follow and 0.4~m/s for crash recovery. Experiments were run in a variety of plant growth stages and weather conditions. The majority of corn rows we tested on were of length 200~m or less, which the robot almost always completed autonomously. At the end of each row, the robot was manually turned to enter another lane, where it then resumed autonomous navigation. 

\subsection{Field Experiment Results}

Due to the limited length of corn rows, a full estimate of the distance between interventions is provided by adding together the lengths of sequential autonomous runs. An autonomous run is considered to have stopped when an intervention occurred outside of reaching the end of the row and manual execution of lane turning. By this metric, we achieved an average of 767 meters between interventions across the three robots we evaluated, and a best-case scenario of 3571 meters before an intervention. These stats were generated across 25315 meters of autonomy.

\begin{figure}[t!]
\centering
\includegraphics[width=8.5cm]{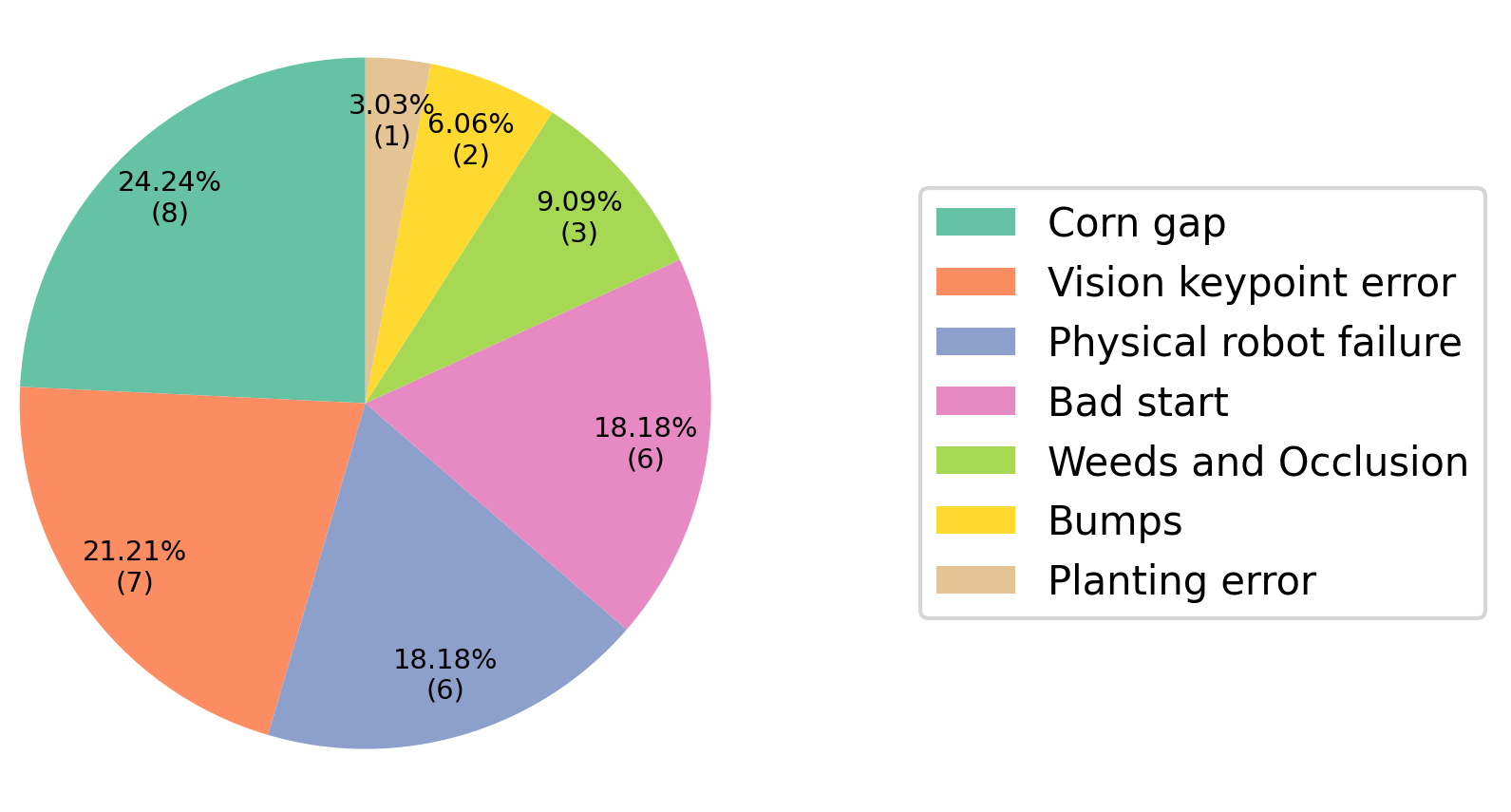}
\centering
\caption{We show the distribution of various causes of \ccr autonomy interventions.}
\label{fig:failure_mode_summary}
\end{figure}

\begin{figure*}[ht!]
    \centering
    \includegraphics[trim=10 0 40 0,clip,width=1\textwidth]{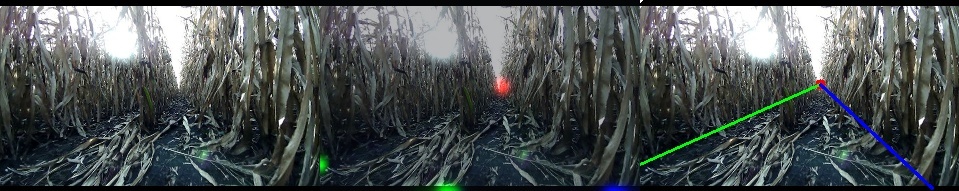}
    \label{error_by_occlusion}
    \includegraphics[trim=10 0 40 0,clip,width=1\textwidth]{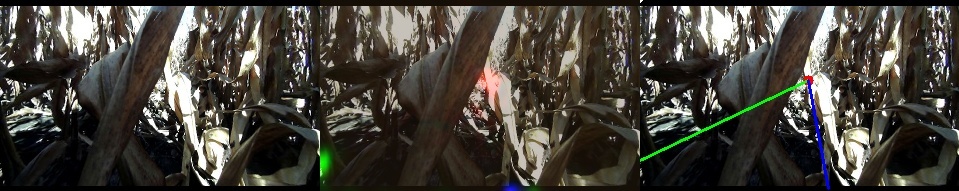}
    \label{error_by_weed}
    \includegraphics[trim=10 0 40 0,clip,width=1\textwidth]{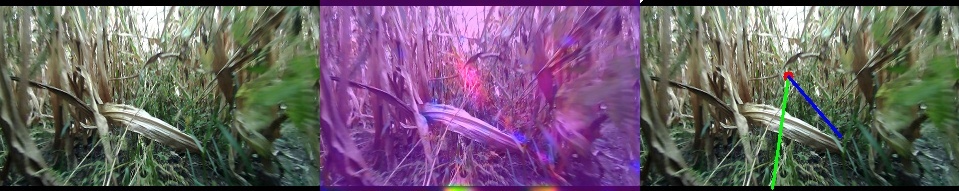}
    \caption{\textbf{Examples of challenging environmental conditions}. \textbf{Top: } An example of a corn gap that led to an autonomy intervention. The keypoint network is confused by the gap and mistakes the neighboring row on the left for the current left row. \textbf{Center: } An example of an occlusion event that led to an autonomy intervention. None of the true keypoints are visible in the image. \textbf{Bottom: } An example of weeds covering the soil which led to an autonomy intervention. The keypoint network is unable to determine the keypoints.}
\label{fig:autonomy_failure_modes}
\end{figure*}

\subsection{Analyzing Failure Modes}

We analyzed the interventions and established a classification schema for causes of autonomy failures. This classification schema consists of the following failure modes. 

\begin{itemize}
    \item \textbf{Vision Keypoint Error} - These failures were caused by errors in our autonomy algorithm during relatively normal conditions.
    \item \textbf{Physical Robot Failure} - These failures were caused by physical failures in the robot, such as motor failure or excessive mud accumulation on the wheels.
    \item \textbf{Corn gap} - These failures were caused by sudden long gaps in the corn on one or both sides. This typically confused the probability map for at least one keypoint, leading the algorithm to be confused as to which row was the true left or right row.
    \item \textbf{Bad start} - These failures were caused by a severely suboptimal initial pose of the robot relative to the corn rows. The pose of the robot was too severe to be corrected even with perfect perception.
    \item \textbf{Weeds and Occlusion} - These failures were caused by objects such as weeds physically blocking the robot, or occluding the cameras. Occlusion would make it impossible to directly perceive one or more of the keypoints, confusing the network. In addition to causing occlusion, the weeds would also sometimes physically impede the robot.
    \item \textbf{Bumps} - These failures were caused by bumpy terrain and ground obstacles that significantly jolted the robot. In these cases, typically the heading of the robot spiked in one direction, making it difficult for the robot to recover.
    \item \textbf{Planting error} - These failures were caused by plants that were erroneously planted in the middle of the gap, instead of in the plant row lines. This is unusual and unexpected behavior in corn. Planting errors would both confuse the vision keypoint model and act as a physical barrier that the robot struggled to overpower.
\end{itemize}

Examples of some of these failures are shown in Figure \ref{fig:autonomy_failure_modes}. A summary of our failures according to this schema is shown in Figure \ref{fig:failure_mode_summary}

\subsection{Key Takeaways from \ccr Deployment}

\begin{itemize}
    
\item  \xxx shows promise as a key part of solving full-field under-canopy autonomy in agriculture. Our experiments show that row follow is very nearly solved in normal situations, and that achieving full-field autonomy would require improving robot hardware and enabling CropFollow++ to handle anomalies encountered due to domain shift such as gaps in the crop rows, presence of weeds, occlusion, and planting errors. Future work could focus on developing a semi-supervised offline learning and self-supervised online learning system to tackle domain shift.

\item  Furthermore, the vision keypoint method provided interpretability of success and failures. Visualizations of the keypoint probability maps are key to deducing how and why various anomalies can cause interventions. We have observed qualitatively that the most reliable vision keypoint estimates tend to come from the cameras closest to the center of the row since they are more represented in the training dataset than other viewpoints. This suggests that more images from other viewpoints are needed in the training dataset to improve the robustness of CropFollow++.

\item  Currently, CropFollow++ only considers the argmax values from the heatmaps as keypoint locations to calculate heading and distance. Other sophisticated heuristics that look at all the clusters in the heatmap would help improve performance in cases such as in the top row of Fig \ref{fig:autonomy_failure_modes}. Also, currently \ccr pipeline fused predictions from multiple cameras with a simple average. Incorporating the variation of heatmap predictions from each camera while fusing the predictions could also be helpful.

\item  We found that correcting for the robot's roll was among the most important refinements we added. The robot's estimation of side distance is sensitive to robot angles, particularly the roll parameter. Adding in this real-time correction was critical to securing performance on rows with non-level ground.

\item We also found that precise calibration was essential to successful row follow, particularly the estimation of the principal point. Our cameras were cheap and did not come with pre-computed calibration, requiring us to find a separate solution. Any error in the principal point would appear as a skew in heading estimation and could lead to frequent crashes.

\item  The results we achieved for CCR autonomy could not have been achieved without the implementation of crash detection and recovery modules. Many times we would crash due to an anomaly or vision keypoint error, and then recover and resume without a problem, having success the second time we encountered the anomaly. Row follow paired with crash detection and recovery will be a key part of solving full-field autonomy.

\end{itemize}
\section*{Acknowledgments}

We acknowledge Naveen Uppalapati’s help in coordinating the collaboration on under-canopy cover crop planting.

\bibliographystyle{IEEEtran}
\bibliography{keypoint_nav}

\end{document}